\documentclass[letterpaper, 10 pt, conference]{ieeeconf}  

\IEEEoverridecommandlockouts                              

\usepackage{graphicx}
\usepackage{multirow}
\usepackage{multicol}
\usepackage{comment}
\usepackage{bm}
\usepackage{color}
\usepackage{diagbox}

\usepackage{amsmath, amssymb}

\title{\LARGE \bf
Contact-Rich Manipulation of a Flexible Object\\based on Deep Predictive Learning using Vision and Tactility
}

\author{ Hideyuki Ichiwara$^{1}$, Hiroshi Ito$^{1,2}$, Kenjiro Yamamoto$^{1}$, Hiroki Mori$^{2}$ and Tetsuya Ogata$^{2}$
\thanks{$^{1}$Hideyuki Ichiwara, Hiroshi Ito and Kenjiro Yamamoto are with Center for Technology Innovation - Mechanical Engineering, Research \& Development Group, Hitachi, Ltd., Ibaraki, 312-0034, Japan
        {\tt\small hideyuki.ichiwara.bn@hitachi.com}}
\thanks{$^{2}$Hiroshi Ito, Hiroki Mori and Tetsuya Ogata are with Department of Intermedia Art and Science School of Fundamental Science and Engineering, Waseda University, Tokyo, 169-855, Japan
        {\tt\small ogata@waseda.jp}}
}

\begin{document}

\maketitle
\thispagestyle{empty}
\pagestyle{empty}

\begin{abstract}
We achieved contact-rich flexible object manipulation, which was difficult to control with vision alone.
In the unzipping task we chose as a validation task, the gripper grasps the puller, which hides the bag state such as the direction and amount of deformation behind it, making it difficult to obtain information to perform the task by vision alone.
Additionally, the flexible fabric bag state constantly changes during operation, so the robot needs to dynamically respond to the change.
However, the appropriate robot behavior for all bag states is difficult to prepare in advance.
To solve this problem, we developed a model that can perform contact-rich flexible object manipulation by real-time prediction of vision with tactility.
We introduced a point-based attention mechanism for extracting image features, softmax transformation for predicting motions, and convolutional neural network for extracting tactile features.
The results of experiments using a real robot arm revealed that our method can realize motions responding to the deformation of the bag while reducing the load on the zipper. 
Furthermore, using tactility improved the success rate from 56.7\% to 93.3\% compared with vision alone, demonstrating the effectiveness and high performance of our method.
\end{abstract}

\section{Introduction}
People can easily perform actions to manipulate flexible objects, such as unzipping a bag, peeling a banana, or changing clothes.
However, it is very challenging to make robots manipulate these flexible objects.
Flexible object manipulations achieved by robots so far include folding clothes, tying ropes, and folding paper \cite{5509439}\cite{miller2012}\cite{7759647}\cite{DBLP:journals/corr/abs-2103-09402}.
For these tasks, it is important to predict how the object will be deformed by the robot's actions, and these tasks have been accomplished mainly using vision.
However, people perform many tasks on a daily basis that cannot be accomplished using vision alone.
This is because our vision is blocked by our own hands and objects when we work with objects.
Robot need to perform such tasks to expand their application range.
In this research, we aim to achieve contact-rich manipulation of flexible objects, which is difficult to do with vision alone, using a two-fingered robot equipped with two tactile sensors as a means to supplement vision.
We focus on the task of unzipping a fabric bag, which has common problems with these tasks, so as not to lose its generality as a validation task.
Common issues are that occlusion occurs with one's hands and fingers and that it is difficult to predict the state of the target object.

One methods to manipulate flexible objects is the physical model-based approach.
In this approach, a physical model of the flexible object is built, and the object is manipulated by predicting how the object will be deformed by the robot \cite{6651522}\cite{6942666}\cite{JIMENEZ2012154}\cite{power2021keep}.
However, the dynamics is complex, and these models are very expensive to build.
Moreover, even when the models are constructed, ideal trajectories are difficult to define for abstract goals such as "unzipping".

\begin{figure}[t]
 \centering
  \includegraphics[width=8.5cm]{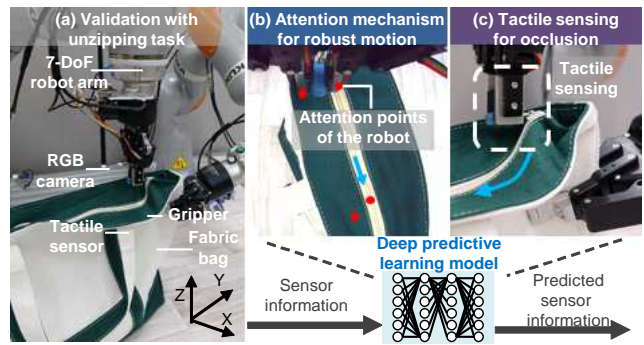}
  \caption{Overview of our study. (a) The experimental setup for flexible fabric bag unzipping.
  (b) Obtain attention points of the robot using point-based attention mechanism. (c) Motion generation based on two tactile sensors attached to the gripper of the robot.}
  \label{fig:experimental_condition}
\end{figure}

Therefore, methods based on deep learning with vision have been investigated for flexible object manipulation.
Deep predictive learning, which learns to simultaneously predict motions and images to generate appropriate robot motions while predicting the behavior of flexible objects
 \cite{DBLP:journals/corr/abs-2103-09402}\cite{yang2016repeatable}\cite{kawaharazuka2019dynamic}, and reinforcement learning-based methods using real robots \cite{tanaka2018emd}\cite{wu2019learning} have been proposed.
These methods are promising for contact-rich flexible object manipulation because they do not require a physical model of the target flexible object since they learn from the robot's motion data.
However, since these methods generate motions based mainly on images, it was difficult to deal with background changes and occlusions that were not present during training.

General-purpose object manipulation methods using deep learning have also been proposed, which are supervised learning-based methods with high sampling efficiency using attention mechanisms \cite{DBLP:journals/corr/LevineFDA15}\cite{ichiwara2021spatial}.
Such methods are expected to enable flexible object manipulation without the need for a large amount of training data, by paying attention to the part of the flexible object's state with complex dynamics that is important for the task.
In addition, Ichiwara et al. \cite{ichiwara2021spatial} showed that narrowing down the image regions that are important for the task and obtaining attention based on real-time prediction of images during the task makes the system robust to background changes and other factors.

Moreover, various types of tactile sensors have been developed to improve the manipulation capability of robots \cite{6583342}\cite{yamaguchi2016combining}\cite{tomo2017covering}\cite{johnson2009retrographic}.
However, most research using these tactile sensors has been done on manipulation by robot hands, including object and material recognition, posture estimation during object grasping, and in-hand manipulation \cite{6907861}\cite{8624961}\cite{tian2019manipulation}.
On the other hand, there are few examples of applications to systems including a robot arm, and they have been applied to simple tasks using reinforcement learning or verified by simulation \cite{lee2019making}\cite{7759578}\cite{vulin2021improved}.

In this work, we propose a deep predictive learning-based robot control system that performs contact-rich flexible object manipulation, which is difficult to achieve with vision alone.
The contributions of this research are as follows:
(1) To make the system robust to dynamic visual changes and occlusions caused by flexible object manipulation, we proposed an attention mechanism to narrow down the important regions of the image.
(2) Using tactile sensors to compensate for occlusion, we confirmed that the success rate of the unzipping task increased from 56.7 to 93.3 \% compared to the case using only vision, demonstrating the effectiveness of tactility in action generation.
(3) We proposed a method that integrates vision, tactility, and motion to generate motions using a deep predictive learning model based on real-time state prediction.

\section{RELATED WORK}

\subsection{Deep Learning-Based Motion Learning}
Deep learning is a powerful machine learning method that has been successfully applied in a wide range of fields such as object recognition and language processing \cite{redmon2018yolov3}\cite{devlin2018bert}, and is also being applied in robot behavior generation.

Yang et al. \cite{yang2016repeatable} used teleoperation to teach a robot to perform a motion and enabled the robot to fold flexible cloth, which was difficult to achieve in the past.
To generate an appropriate robot motion while predicting the behavior of a flexible object, deep predictive learning has been proposed, in which learning is performed to predict motion and an image at the same time based on training data.
Suzuki et al. \cite{DBLP:journals/corr/abs-2103-09402} enabled  robots to manipulate ropes by two arms on the bases of the method of Yang et al. \cite{yang2016repeatable}.
In addition to the image and the joint angle of the robot, the proximity sensor was used to recognize the rope state.
However, since the proximity sensor was used to recognize whether or not there is a rope in the robot hand, high-dimensional contact information such as tactility was not used. 

In addition, sampling efficient methods \cite{DBLP:journals/corr/LevineFDA15}\cite{ichiwara2021spatial}\cite{finn2016deep} have been proposed using the attention mechanism.
Levine et al. \cite{DBLP:journals/corr/LevineFDA15}\cite{finn2016deep} proposed using CNN and soft argmax to obtain the position coordinates in the image for image feature extraction.
Since positional information is important for the robot's task, the sampling efficiency was improved by compressing the information-rich image to a lower order.
Ichiwara et al. \cite{ichiwara2021spatial} proposed a new method based on the method of Levine et al. \cite{DBLP:journals/corr/LevineFDA15}\cite{finn2016deep} that was made robust to changes in background and object positions by predicting attention points on the basis of image prediction. 
By adding top-down attention based on prediction in addition to bottom-up attention from images, it is possible to extract only the location information important for the task.
These methods using the attention mechanism have been validated for tasks such as object grasping and pick-and-place, where the object has a fixed shape.
On the other hand, they have not been validated for flexible object manipulation, which involves dynamic vision changes due to object manipulation.

\subsection{Manipulation using Tactility}
Much of the research on manipulation using tactility has been done on in-hand manipulation with robot hands \cite{6907861}\cite{8624961}\cite{tian2019manipulation}.
As for the application to systems that include a robot arm, a robot manipulation method based on deep reinforcement learning using tactility has been proposed, and using tactility has been shown to improve search efficiency \cite{vulin2021improved}.
However, the effectiveness of this method has not yet been verified on a real robot.
In research using real robots, a reinforcement learning-based method using vision and tactility has been proposed, and using tactility has been shown to enable stable learning \cite{7759578}.
This method has been validated with a simple task such as using a five-axis robot to control a single finger attached to a tactile sensor to make a pole rotating in two axes stand upright.

In this study, we propose a deep predictive learning model that integrates and predicts vision, tactility, and motion, using attention mechanisms.
Furthermore, as an example of contact-rich flexible object manipulation, we test the effectiveness of our model on a real robot for the unzipping task of a fabric bag.
Finally, we show that the tactility is important in the task.

\section{METHOD}
\begin{figure*}[t]
 \centering
  \includegraphics[width=18.0cm]{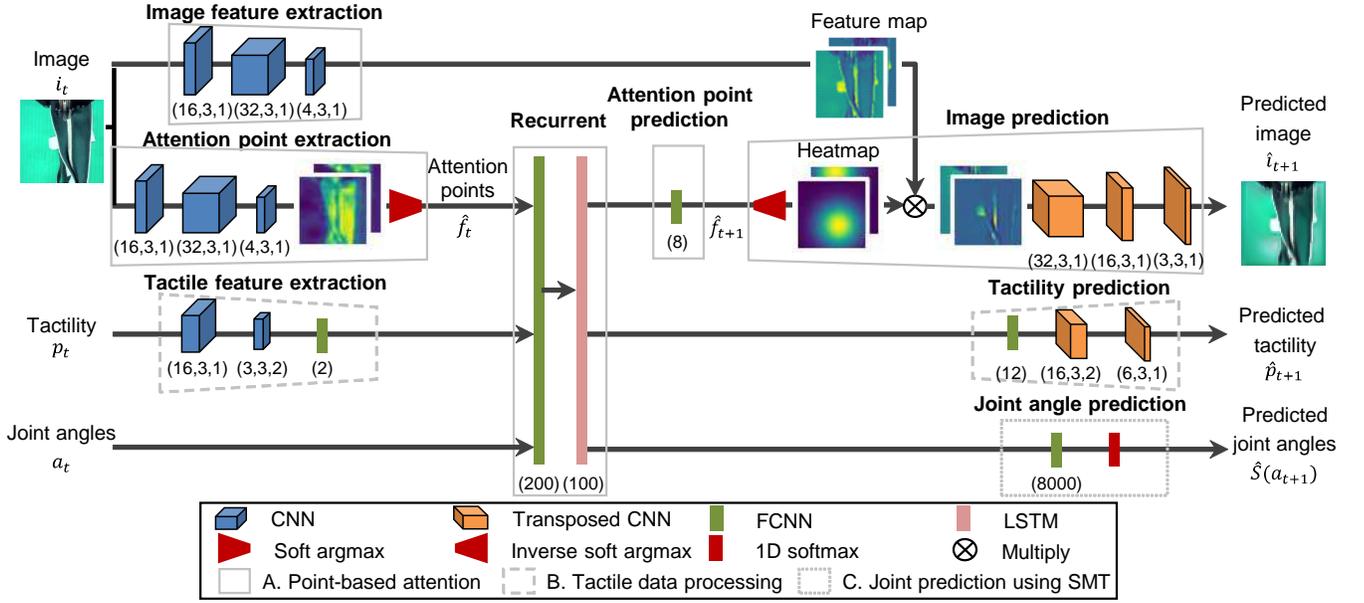}
  \caption{The proposed model architecture for contact-rich flexible object manipulation using tactility. It consists of point-based attention, which extracts image features and position information, tactility data processing, and joint prediction. 
  }
  \label{fig:model}
\end{figure*}
\begin{figure}[t]
 \centering
  \includegraphics[width=8.0cm]{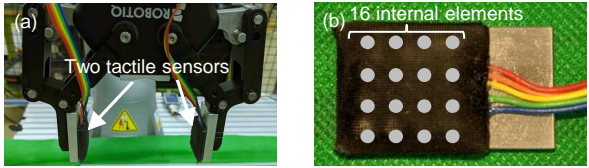}
  \caption{(a) Two-finger gripper with two tactile sensors. (b) Tactile sensor that has 16 built-in elements.}
  \label{fig:experimental_condition2}
\end{figure}

Fig. \ref{fig:model} shows the proposed model for contact-rich flexible object manipulation using tactility.
The network parameters in this study are shown in Fig. \ref{fig:model}. For example, the explanation in CNN and Transposed CNN shows (channels, kernel size, strides), and the explanation in FCNN and LSTM shows (number of nodes).
The camera image $\bm{i}_t$, tactility data $\bm{p}_t$, and joint angle $\bm{a}_t$ of the robot at time $t$ are input to the model, which predicts the camera image $\hat{\bm{i}}_{t+1}$, tactility data $\hat{\bm{p}}_{t+1}$, and joint angle $\hat{S}(a_{t+1})$ after softmax transformation (SMT) at the next time $t+1$.
Here, it is important to simultaneously predict sensory inputs (images, tactility) other than motion. 
If the sensory inputs are not predicted, the sensory input are not referenced and only the time $t$ and $t+1$ motions are learned, and the appropriate motion is not generated \cite{ichiwara2021spatial}.

To train the model, we used the time series data when the robot executes the action by human operation as the training data. 
The model was then trained to output the data at the next time $t+1$ from the data at time $t$. 
When the robot executes a motion using the learned model, each data at the current time is input to the model, and the predicted joint angle data is input to the robot as commands.
Our model has the following features.
To deal with dynamic visual changes, we used the point-based attention mechanism on the basis of prediction, as shown by the solid line in Fig. \ref{fig:model}.
To make it robust to tactility position shifts caused by shifting of the puller, we processed the tactility data using CNNs, as shown by the dashed lines. 
To generate finer movements of the robot, SMT was introduced, which is a method of expressing joint angles in a sparse form, as shown by the dashed lines. 

\subsection{Point-Based Attention Mechanism}
The key components of the attention mechanism are the blocks shown by the solid gray lines in Fig. \ref{fig:model}.
The attention point extraction block using CNNs and soft argmax outputs the positional coordinates of points in the image (called the attention point in this paper) from the input image.
Here, soft argmax obtains the index that takes the maximum value of the array and relaxes the conditions of the argmax function, which has difficulty back-propagating errors \cite{luvizon20182d}.
In this block, the feature maps of the input image are obtained using CNNs. 
In addition, soft argmax is used to extract high-intensity positions in the feature map and treat them as attention points. 
Next, the recurrent block, attention points prediction block, and joint angle prediction block using the FC layer and LSTM layer predict the attention points, joint angle and tactility feature at the next time $t+1$ from those data at time $t$.
In addition, the image prediction block predicts the image at the next time $t+1$ on the basis of the prediction attention points and the feature maps obtained from the image feature extraction block.
Specifically, it generates heat maps with high intensity at the prediction points of interest and predicts the image at the next time $t+1$ on the basis of the feature maps weighted by the heat maps.
This is because the heat map alone is not sufficient to predict the image, but the image can be easily predicted by using the information near the attention points among the image features obtained from the image feature extraction block.

Note that since all blocks are end-to-end connected and the attention points are determined by a self-organized manner, the training data of the attention points is not necessary. 
The points are obtained in such a way as to minimize the prediction error between the image and the joint angle, and they are obtained at positions important to the task.
The predicted attention points output by attention point prediction block are called predicted attention points because they are obtained to minimize the prediction error of the image.

\subsection{Tactility Data Processing using CNNs}

In this study, we used the tactile sensor (XR1944 from XELA Robotics) attached to the gripper shown in Fig. \ref{fig:experimental_condition2}.
This sensor has 16 elements inside and uses magnetism.
For each contact point, the digital values in the three axial directions of the pressure direction and the shear direction can be detected, and one sensor outputs 4$\times$4$\times$3 values. Shear force and normal force resolution are 0.1 gf and 1 gf.
Therefore, the output of the sensor can be treated in the same way as the image, and as shown by the broken line in Fig. \ref{fig:model}, CNNs were used for extracting tactile features. 
In image processing, CNN usage is robust to positional shifts \cite{ngiam2010tiled}.
In this study, we used CNNs for tactile feature extraction to make our system robust to positional shifts of the puller.
In addition, for some tasks, it may be better to have a higher sensitivity to position shift.
In this case, we can increase the sensitivity by reducing the CNN stride, increasing the number of tactile dimension input to the recurrent block, etc.

\subsection{Joint Prediction using Softmax Transformation(SMT)}
To enhance the learning, SMT \cite{8329798} was used to predict joint angles.
SMT converts joint angles into sparse representations instead of representing them as continuous values such as radians.
For example, one joint angle $a$, which can take values from 0 to 1, is transformed using $J$ neurons:
\begin{eqnarray}
  n_j^a=\frac{\exp(\frac{-\|\frac{j}{J}-a\|^2}{\sigma})}{\sum_{j\in J}\exp(\frac{-\|\frac{j}{J}-a\|^2}{\sigma})},  \label{eq:SMT} 
\end{eqnarray}
where $n_j^a$ represents $j$th neuron, and $\sigma$ is a parameter that determines the distribution after the transformation.
In our experiment, $J$ was set to 1000, and $\sigma$ was set to 0.1.
On the other hand, the joint angle prediction block shown in Fig. \ref{fig:model} predicts the joint angle after SMT through a 1-dimensional (1D) softmax layer represented by $J$ neurons per joint angle.
When testing after training, the output of the model represented by SMT was converted into the joint commands of the robot:
\begin{eqnarray}
  a= \sum_{j\in J}n_j^a \times \frac{j}{J}. \label{eq:iSMT} 
\end{eqnarray}
In this way, by using multidimensional neurons for each joint, finer movements can be generated. 

\subsection{Loss Function}

The loss function was defined as follows:
\begin{eqnarray}
  g&=&\sum_{t\in T}\{g_{i}(\bm{i}_{t+1},{\hat{\bm{i}}}_{t+1}) + g_{p}(\bm{p}_{t+1},{\hat{\bm{p}}}_{t+1})+  \nonumber \\
  &&g_{a}(S(\bm{a}_{t+1}),{\hat{S}}(\bm{a}_{t+1}))+\alpha g_{f}(\hat{\bm{f}_{t}},{ \hat{\bm{f}}}_{t+1})\}, \label{eq:TotalLoss} \\
  g_{i}&=&\frac{1}{H_i\times W_i\times C_i}\|\hat{\bm{i}}_{t+1} - \bm{i}_{t+1} \|_2^2, \label{eq:ImgLoss} \\
  g_{p}&=&\frac{1}{H_p\times W_p\times C_p}\|\hat{\bm{p}}_{t+1} - \bm{p}_{t+1} \|_2^2, \label{eq:TacLoss} \\
  g_{a}&=&\frac{1}{M\times J} \sum_{l\in M}\sum_{m\in J} S(\bm{a}_{t+1})_l^m \log(\hat{S}(\bm{a}_{t+1})_l^m),    \label{eq:AngleLoss} \\
  g_{f}&=&\frac{1}{K}\| \hat{\bm{f}}_{t} - \hat{\bm{f}}_{t+1}\|_2^2, \label{eq:FeatureLoss} 
\end{eqnarray}
The sequence length of the training data is $T$, the time is $t$, the image is $\bm{i}\in\mathbb{R}^{H_i\times W_i\times C_i}$, the tactility is $\bm{p}\in\mathbb{R}^{H_p\times W_p\times C_p}$, the joint angle $\bm{a}\in\mathbb{R}^M$ after SMT is $S(\bm{a})\in\mathbb{R}^{M\times J}$, and the coordinates of the attention points are $\bm{f}\in\mathbb{R}^{K}$.
$g_i$ and $g_p$ are prediction errors of the image and tactility using the mean square error.
$g_a$ is the prediction error of the joint angle after SMT using cross-entropy loss.
Also, $g_f$ denotes the error in the Euclidean distance between the attention points $\hat{f_{t}}$ of the encoder output and the attention points $\hat{f}_{t+1}$ of the decoder input. 
$\alpha$ denotes the weight of the loss function.
By adding $g_f$, the predicted attention points was efficiently searched near the current attention points.
This is because the attention points are obtained for the gripper and the target object, and the position does not change significantly with a one-step time change.
In this study, images were RGB color images with a resolution of $64\times64$, so $H_i=64, W_i=64, C_i=3$. One tactile sensor consists of 4×4 elements and outputs values in the 3-axis direction, so $H_p=4, W_p=4, C_p=6$.
The robot arm used has seven axes and the end-effector was a two-fingered gripper, so $M=8$ and $J=1000$.
The number of the attention points was four, and considering the two coordinates x and y, $K=2\times4$. 
Also, $\alpha=0.0001$ (epoch=0), $\alpha=0.1$ (epoch$>$1000), and $\alpha$ was increased in steps.

\section{EXPERIMENTS}
\begin{figure}[t]
  \vspace{0.2cm}
  \begin{minipage}[t]{\hsize}
    \centering
    \includegraphics[width=8.0cm]{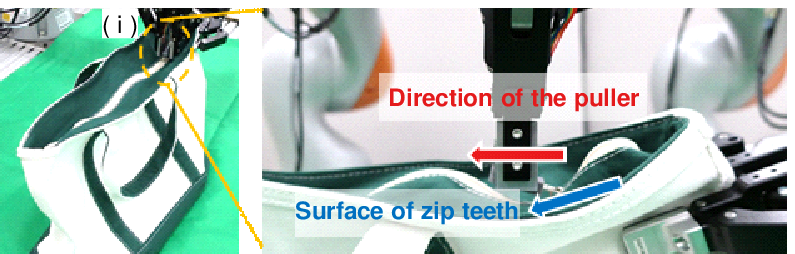}
  \end{minipage}
  \begin{minipage}[t]{\hsize}
    \centering
    \includegraphics[width=8.0cm]{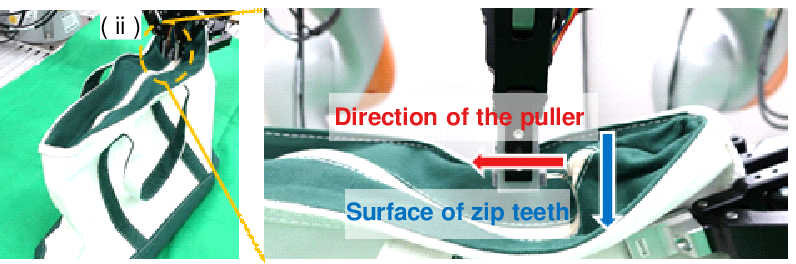}
  \end{minipage}
  \begin{minipage}[t]{\hsize}
    \centering
    \includegraphics[width=8.0cm]{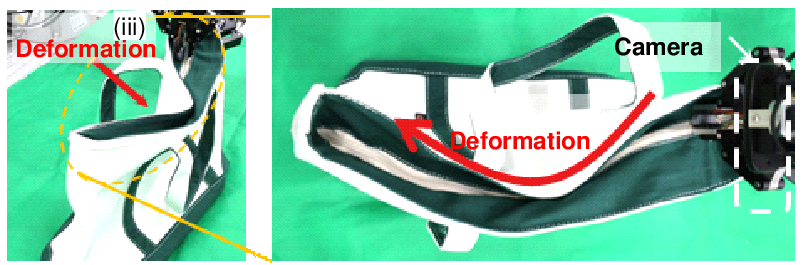}
  \end{minipage}
  \caption{The initial bag state to verify the task success rate. $\rm(\hspace{.18em}i\hspace{.18em})$: without visible bag deformation. $\rm(\hspace{.08em}ii\hspace{.08em})$: without visible bag, but the puller is perpendicular to the zip teeth. $\rm(i\hspace{-.08em}i\hspace{-.08em}i)$: with visible bag deformation.}
  \label{fig:ExperimentalPattern}
\end{figure}
\subsection{Hardware}
The experimental environment is shown in Fig. \ref{fig:experimental_condition}(a) and Fig. \ref{fig:experimental_condition2}.
As shown in Fig. \ref{fig:experimental_condition}(a), it consists of two robot arms of KUKA LBR iiwa 14 R820. 
Robotiq 2F-85 Adaptive Gripper was attached to each arm. 
An RGB camera was placed on the gripper, and Buffalo's BSW500M Series was used. 
In addition, tactile sensors in Fig. \ref{fig:experimental_condition2}(b) were attached to each of the two fingers of the gripper, as shown in Fig. \ref{fig:experimental_condition2}(a). 
For the fabric bag, a commercially available tote bag manufactured by Captain Stag was used. 
The puller was made with a 3D printer so that it could be easily grasped with the gripper. 
In this study, we focused on performing tasks in situations where the task state cannot be recognized only by images. 
Therefore, the upper part of the bag was fixed with one arm, the puller was grasped by the arm to be controlled, and the state of contact with the object was set as the initial state of the task.
At this time, since the initial state of the arm was the same, the position of the bag was placed within the range where the puller could be grasped.
The bag and the puller were not fixed in a strict position, but were set to be in approximately the same position.
Then, the puller was pulled to the other side of the bag, and the state where the bag was opened, which was defined as the end of the task. 

\subsection{Evaluation}
The purpose of the experiment was to verify the effectiveness of our model.
In particular, we investigated the effectiveness of using CNNs for tactility feature extraction, SMT, and tactility.
Therefore, we compare the success rates of the tasks in the five models.

\begin{itemize}
 \item[(A)] Without SMT and tactility, same as \cite{ichiwara2021spatial} 
 \item[(B)] With SMT without tactility
 \item[(C)] With tactility without SMT
 \item[(D)] With SMT and tactility using CNN for tactility (ours)
 \item[(E)] With SMT and tactility using FC for tactility
\end{itemize}

Fig.\ref{fig:ExperimentalPattern} shows the initial bag state to verify the task success rate.
$\rm(\hspace{.18em}i\hspace{.18em})$ was when the bag was not deformed significantly as shown in the figure on the left.
Also, as shown in the figure on the right, the direction of the puller was almost horizontal to the zip teeth, and pulling the puller in the direction of the teeth end opens the bag, so it was not expected to be difficult to open.
$\rm(\hspace{.08em}ii\hspace{.08em})$ had no visible deformation in the figure like $\rm(\hspace{.18em}i\hspace{.18em})$.
However, as shown in the figure on the right, unlike $\rm(\hspace{.18em}i\hspace{.18em})$, the puller is perpendicular to the teeth, so if it is pulled directly in the direction of the teeth end, it will become stuck and the zipper cannot be opened.
Furthermore, the camera was located above the zipper, a case that we expected to require particularly tactility, since it appeared to be able to be opened by pulling it straight out using only vision.
$\rm(i\hspace{-.08em}i\hspace{-.08em}i)$ was a case where the bag was given a visible deformation, as shown in the left figure, and the arm fingertips needed to be moved appropriately to match the direction of the zipper.
The deformation was also visible from the direction of the camera, as shown in the right figure.
For each case, we performed 30 trials each and evaluated the success rate.
The trial time was set to 10 seconds, which is the same as the training data, and if it did not finish within the time, it was defined as a failure. 

\subsection{Training Setup}
As for the training data, the bag was randomly deformed, and a human operated the robot to acquire 36 patterns. 
Generally, Robot operation methods for acquiring the training data include arm hand position / posture control method using an operation user interface such as a game controller, bilateral control method \cite{9344611}, and direct teaching method.
In this research, we chose the operation by direct teaching from the viewpoint of ease of implementation and the fact that the operator can operate while feeling the reaction force.
The operator grabbed the arm and performed the unzipping motion of the bag while keeping the load on the hand small.
Joint angle data of the robot arm, image data of the camera, and tactile sensor data were acquired for 10 seconds and 20 Hz.
Each model was trained at 10,000 epochs and did not use pre-trained models.
The batch size was 18, and the input data was scaled to [0,1.0].
The Adam optimizer \cite{kingma2014adam} was used for training.
Optimizer parameters were set to $\alpha=0.001$, $\beta_1 = 0.9$, and $\beta_2 = 0.999$.
We used TensorFlow as a software library.
To train the models, the ABCI system of AIST was used.
Due to the large network size of our model, large computational resources 
with graphics processing unit (GPU) memory were required for training.
The ABCI system has 16GB per node, and we used two nodes in this study.

\section{RESULTS}
\begin{figure*}[t]
 \centering
  \includegraphics[width=17.0cm]{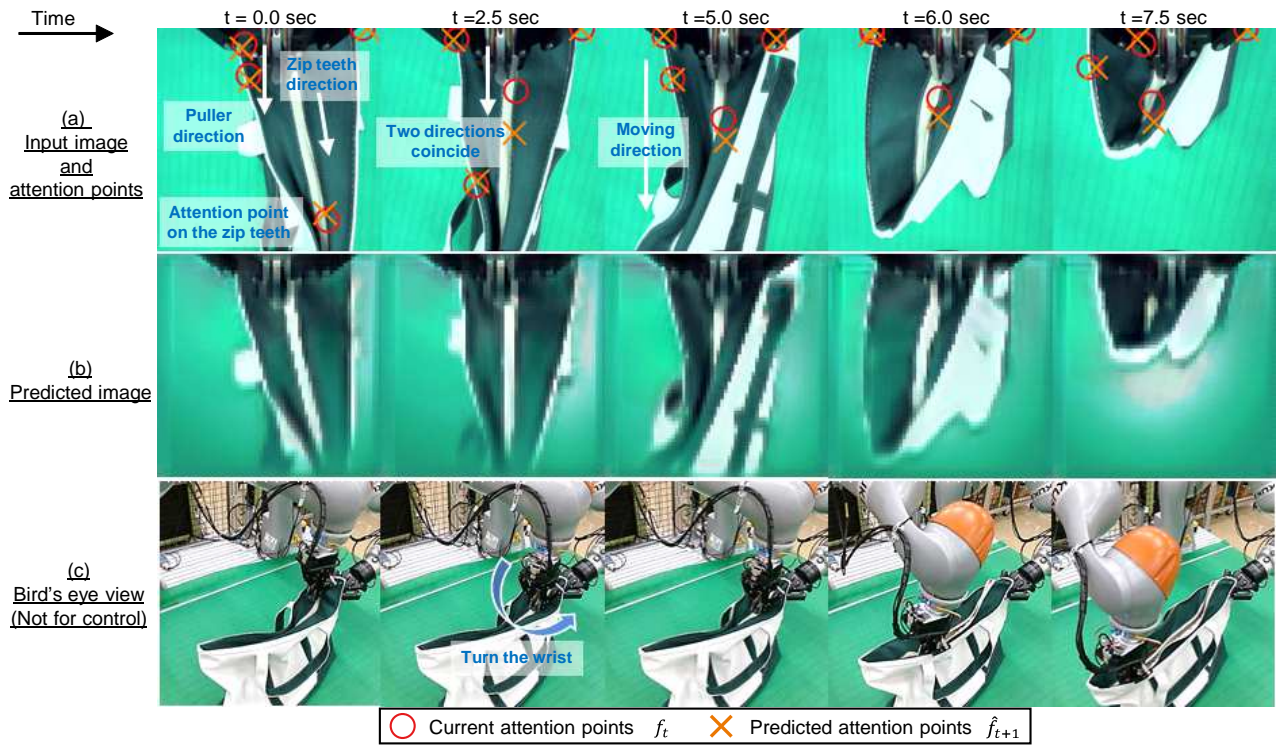}
  \caption{Snapshots of unzipping task with our model in the initial bag state  $\rm(i\hspace{-.08em}i\hspace{-.08em}i)$.
  (a): images of the robot and attention points. The attention points are the position to which the robot was paying attention . (b): predicted images of time $t+1$. (c): bird's eye view camera images. 
  }
  \label{fig:snapshot_propose}
\end{figure*}

\begin{table}[t]
\vspace{0.2cm}
\begin{center}
 \caption{Task success rate for each model and initial bag state.}
  \begin{tabular}{l|ccc|c} \hline
    \diagbox{Model}{Bag state} & $\rm(\hspace{.18em}i\hspace{.18em})$ & $\rm(\hspace{.08em}ii\hspace{.08em})$ & $\rm(i\hspace{-.08em}i\hspace{-.08em}i)$ & All \\ \hline\hline
    (A) Base \cite{ichiwara2021spatial}&23.3\%&0.0\%&26.7\%&16.7\% \\ \hline
    (B) w/ SMT&83.3\%&0.0\%&86.7\%&56.7\% \\ \hline
    (C) w/ tac. (CNN)&63.3\%&23.3\%&36.7\%&41.1\% \\ \hline
    {\bf (D) w/ SMT, tac. (CNN)}&{\bf 96.7\%}&{\bf 86.7\%}&{\bf 96.7\%}&{\bf 93.3\%} \\ \hline
    (E) w/ SMT, tac. (FC)&63.3\%&63.3\%&56.7\%&61.1\% \\ \hline 
  \end{tabular}
  \label{tab:SucessRate}
\end{center}
\end{table}

\begin{table}[t]
\centering
\caption{Load on the fingertips in each model and initial bag state.}
\begin{tabular}{c||cc||cc} \hline
\multirow{2}{*}{\diagbox{Model}{Bag state}} & \multicolumn{2}{c||}{Maximum {[}N{]}} & \multicolumn{2}{c}{Average {[}N{]}} \\
                                            & $\rm(\hspace{.18em}i\hspace{.18em})$ & $\rm(i\hspace{-.08em}i\hspace{-.08em}i)$ & $\rm(\hspace{.18em}i\hspace{.18em})$ & $\rm(i\hspace{-.08em}i\hspace{-.08em}i)$ \\ \hline  \hline
(B) w/o tactility                             & 15.8             & 22.5             & 8.0              & 10.1              \\ \hline
{\bf (D) w/ tactility}                        & 12.5             & {\bf 14.7}       & 7.8              & {\bf 8.0}        \\ \hline           
\end{tabular}
\label{tab:EndeffectorForce}
\end{table}

Table \ref{tab:SucessRate} shows the task success rates for each model and initial bag state.
The comparison of the success rates of each model revealed the following. 
\begin{itemize}
    \item (A) and (B), (C) and (D) differed only in the presence or absence of SMT.
    In the comparison, the success rates of (B) and (D) using SMT were higher, indicating the effectiveness of SMT.
    \item (B) and (D) differed only in the presence or absence of tactility. 
    In the comparison, the success rate of the model using tactility was higher, which indicates the effectiveness of tactility usage.
    \item (D) and (E) differed only in the tactility data processing method.
    In the comparison, the model that extracted features by CNNs had a higher success rate than the model that extracted features by FC, which indicates the effectiveness of CNNs for tactility data processing.
\end{itemize}
From the above, we were able to verify the effectiveness of using CNNs for tactility data processing, SMT, and tactility.
Furthermore, we were able to achieve the highest success rate of 93.3\% for our model with the three methods, demonstrating the effectiveness of our model.
In addition, two questions were investigated on the basis of the results.

\subsection{Is the attention mechanism functioning properly?}
Fig. \ref{fig:snapshot_propose} shows (a) the input image and attention points, (b) predicted image, and (c) bird's eye view for our model (D) in the initial bag state  $\rm(i\hspace{-.08em}i\hspace{-.08em}i)$, where the visible deformation of the bag is large and the effect of attention can be easily determined.
Time elapsed with the image on the right.
There are two types of attention points: $\bigcirc$ indicates current attention points, and × indicates predicted attention points.
Some attention points were on the gripper and zip teeth. 
These are important positions for the task. 
In addition, the attention points obtained on the zip teeth were ahead of the current attention points in the direction of motion, and the predicted attention points were also appropriately predicted.
From the above, it can be considered that the attention points are obtained appropriately.
Note that similar attention points were found for the other bag initial state.
In addition, at the initial time 0 sec, the directions of the puller and the zip teeth were misaligned, but the attention points were output on the zip teeth, and at 2.5 sec, the arm wrist was rotated to match the direction of the puller and the zip teeth.
After that, the zipper was successfully opened to the end by generating arm movements on the basis of the attention points obtained on the zip teeth.
From the above, it can be concluded that the attention mechanism was functioning properly, as the unzipping task was successfully performed by generating motions on the basis of the obtained attention points.



\subsection{Does tactility contribute to action generation?}
Table \ref{tab:EndeffectorForce} shows the maximum and average loads on the arm fingertips at the time of success in bag states $\rm(\hspace{.18em}i\hspace{.18em})$ and $\rm(i\hspace{-.08em}i\hspace{-.08em}i)$, where the success rates were nearly equal for models (B) and (D) without and with tactility.
Here, the maximum load was the maximum value of the load for each trial averaged over the number of trials, and the average load was the average of all trials.
Both the maximum and average loads were smaller when using tactility in all bag conditions.
In particular, in the initial bag state $\rm(i\hspace{-.08em}i\hspace{-.08em}i)$, where the deformation of the bag was large and the operation based on tactility was necessary comparing to $\rm(\hspace{.18em}i\hspace{.18em})$, the difference between the two cases was large.
These results suggest that the motion was generated on the basis of tactility.

\section{CONCLUSION}
We proposed a deep predictive learning robot control system that performs contact-rich manipulation of a flexible object, which was difficult to achieve by vision alone.
Furthermore, we demonstrated its effectiveness with unzipping of a fabric bag. 
Our methods introduced an attention mechanism based on convolutional neural networks (CNNs) and soft argmax for extracting image features, softmax transformation for predicting motions, and CNN for extracting tactile features. 
In the experiment, the robot was able to generate motions while responding to the bag state, which dynamically deforms when it was unzipped, and showed a high success rate.
In addition, the importance of tactility in the task was demonstrated by not only the success rate but also the difference in the load on the zipper. 
In this study, we considered the unzipping task as an example.
Our model is not task-specific but can be widely applied to tasks. 

\section*{Acknowledgement}
AI Bridging Cloud Infrastructure of National Institute of
Advanced Industrial Science and Technology was used.

\bibliographystyle{IEEEtran}
\bibliography{ref}

\end{document}